\documentclass[11pt]{article}

\usepackage[margin=1in]{geometry}
\usepackage[numbers,sort&compress]{natbib}
\usepackage[version=3]{mhchem}
\usepackage{graphicx}
\usepackage{float}
\usepackage[english]{babel}
\usepackage[T1]{fontenc}
\usepackage{hyperref}
\usepackage{multirow}
\usepackage{booktabs}
\usepackage{array}
\usepackage{amsmath}
\usepackage{amssymb}

\begin{document}

\title{A General-Purpose Molecular Foundation Model Transfers Across Diverse Olfactory Tasks}

\author{
Yikun Han$^{1,\ast}$,
Yi Wang$^{1,\ast}$,
Neil Mankodi$^{1}$,
Stephen Yang$^{2}$,
and Ambuj Tewari$^{1,2}$
\\[0.8em]
\small $^{1}$Department of Statistics, University of Michigan, Ann Arbor, Michigan, USA
\\
\small $^{2}$Computer Science and Engineering Division, University of Michigan, Ann Arbor, Michigan, USA
\\[0.4em]
\small $^{\ast}$These authors contributed equally to this work.
}

\date{}

\maketitle

\begin{abstract}
Foundation models have transformed molecular property prediction, yet it remains unclear whether a molecular foundation model, fine-tuned on a single canonical olfactory prediction task, can learn representations that transfer across diverse machine olfaction problems. We investigate this question by fine-tuning Uni-Mol2 on the GS-LF benchmark for multi-label odor descriptor prediction and evaluating the resulting model, without additional deep-learning training, on four complementary downstream settings: cross-dataset odor descriptor prediction, odorous-versus-odorless classification, enantiomer evaluation, and odor mixture discriminability. The fine-tuned model matches or exceeds the performance of the state-of-the-art olfaction-specific baseline on the primary GS-LF benchmark and consistently transfers across these downstream evaluations. The enantiomer analysis further shows that the learned representations yield distinct predictions within each enantiomer pair, although accurately predicting the perceptual consequences of stereochemistry remains an open challenge. Together, these results support a train-once, transfer-across-tasks paradigm for machine olfaction and suggest that chemically pretrained molecular representations provide a strong foundation for transferable olfactory prediction.
\end{abstract}

\section{Introduction}


Deep learning has driven remarkable advances in computer vision as well as speech and natural language processing. Availability of web-scale datasets, high-performance GPUs, and novel neural network architectures have completely transformed these fields. In particular, recent years have seen the emergence of foundation models trained on massive amounts of data that can be adapted to a variety of downstream tasks via fine-tuning or few-shot learning \cite{bommasani2021opportunities,kirillov2023segment}. However, progress in olfaction is more limited compared to vision, audition, and language. A key roadblock is limited data availability which hinders advances in predictive modeling. The disparity between vision, audition, and language on the one hand and olfaction on the other raises an important question: can the powerful techniques such as foundation models that revolutionized other sensory and cognitive domains similarly bridge the gap between machine and human performance in olfaction, enabling robust and highly accurate predictive models?

The enthusiasm about the potential of foundation models in machine olfaction has to be tempered with the fact that olfaction is biologically complex. Humans possess approximately 400 olfactory receptors (ORs) responsible for odor detection, suggesting that odor perception, much like other molecular properties, can be modeled computationally in principle but we are not there yet \cite{trimmer2019genetic}. The task of accurately predict the smell of molecules solely based on their chemical structures has seen a lot of progress in recent years but remains far from fully solved.

The task of linking an odorant's molecular structure to its olfactory perceptual qualities is called quantitative structure-odor relationship (QSOR) modeling. Although QSOR has historically received less attention compared to its more established counterpart, quantitative structure–activity relationships (QSAR), existing research clearly demonstrates the predictability of odor from molecular structure. Nevertheless, data scarcity remains the primary bottleneck, constraining progress in the field of machine olfaction. Early QSOR research relied on limited datasets characterized by a small number of odor descriptors and stimuli. These pioneering efforts typically utilized statistical methods and rudimentary neural networks. These led to innovations at the time but were restricted by data quality and scale \cite{chastrette1996structure, dravnieks1985atlas, arctander1969perfume}.

A major milestone in QSOR research was the DREAM Olfactory Prediction Challenge, marking the transition from purely statistical methods to machine learning-driven approaches \cite{keller2017predicting}. The challenge provided a dataset comprising 21 odor attributes for 476 molecules, with the top-performing model utilizing a random forest algorithm, a traditional yet effective method in classical machine learning, trained on Dragon molecular features and Morgan fingerprints \cite{mauri2006dragon, rogers2010extended}. Despite demonstrating machine learning's potential in QSOR, data limitations continued to restrict further breakthroughs.

Recent QSOR studies have advanced significantly by incorporating graph-based machine learning methods, notably graph neural networks (GNNs). These methods represent molecules as graphs, effectively capturing molecular structures and demonstrating superior performance compared to earlier statistical approaches \cite{gilmer2017neural, kipf2016semi}. A significant breakthrough occurred with the development of the GS-LF dataset, an expert-annotated dataset containing 138 odor descriptors for approximately 5,000 molecules, sourced from GoodScents and Leffingwell databases \cite{goodscents2019, leffingwell2005}. Models trained on GS-LF notably surpassed the performance of average human panelists on 53\% of tested molecules, illustrating the potential of advanced machine learning to predict odor from molecular structures alone.


While QSOR has recently expanded its scope to include more complex tasks such as mixture modeling and perceptual distance estimation between pairs of odor mixtures, these novel branches have primarily focused on enhancing the understanding of complex odor interactions rather than improving single-molecule odor descriptor predictions \cite{kowalewski2021system, sisson2023olfactory, snitz2013predicting, ravia2020measure, dhurandhar2023expansive}. Consequently, progress on the mainstream QSOR task, odor descriptor prediction on benchmark datasets like GS-LF, has stagnated, motivating the exploration of novel strategies to overcome existing limitations.


 Motivated by recent success in adapting a general purpose molecular foundation model to an olfactory task~\citep{wadell2026foundation}, we investigate whether a fine-tuning a foundation model on a single olfactory task builds a representation that transfers successfully to multiple downstream olfactory tasks. Instead of training models from scratch solely on limited QSOR-specific data, we adapt molecular representations learned from extensive chemical pretraining to olfaction by fine-tuning the model on the GS-LF dataset. We show the resulting model's strong transferability across a diverse range of downstream olfactory tasks without or with minimal task-specific training. Overall, our model demonstrates the potential of foundation models to support a scalable train-once, apply-across-tasks paradigm for machine olfaction.

\section{Methodology}
In this section, we present our transfer-learning framework for machine olfaction. We first provide an overview of our approach, then detail our transfer learning methodology using Uni-Mol2 as the  foundation model, and finally describe the fine-tuning and evaluation procedures used to assess its performance and transferability across olfactory tasks.

\subsection{Overview}

\begin{figure}[htbp]
\centering
\includegraphics[width=\textwidth]{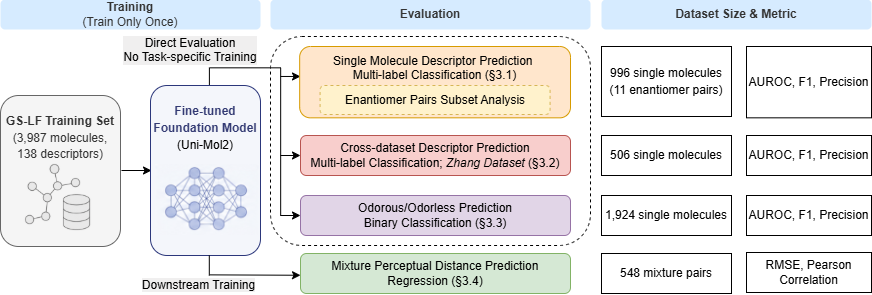}
\caption{
Overview of the modeling pipeline for machine olfaction tasks. The GS-LF dataset is used to fine-tune a molecular foundation model. The resulting model is directly evaluated on three tasks without additional training: (i) single-molecule olfactory label prediction, including an enantiomer-pair subset analysis (§3.1.1); (ii) cross-dataset descriptor prediction on the Zhang dataset (§3.1.2); and (iii) single-molecule odor/odorless binary classification (§3.1.3). For mixture perceptual distance prediction (§3.1.4), the learned molecular representations are used for downstream training.}
\label{fig:overview}
\end{figure}

An overview of our methodological framework is illustrated in Figure~\ref{fig:overview}. First, we fine-tune a molecular foundation model on the GS-LF dataset. The resulting model is then evaluated across four olfactory tasks:  (i) single molecule odor descriptor prediction on the GS-LF test set, with enantiomer pairs olfactory label prediction as a particularly challenging subtask; (ii) cross-dataset odor descriptor prediction on the Zhang et al. dataset~\citep{zhang2024deep}; (iii) single-molecule odorous/odorless classification; and (iv) mixture perceptual distance prediction. These tasks correspond to different machine learning problems ranging from multilabel classification to binary classification to regression. We therefore use a variety of evaluation metrics appropriate to the task at hand, ranging from standard classification metrics (AUROC, F1, Precision) to regression metrics (RMSE, Pearson correlation). The dataset sizes range from a few hundreds to a couple of thousands reflecting the small sizes of even the largest publicly available olfaction datasets.

Together, these evaluations assess whether the model trained on a single olfactory dataset can generalize across distinct olfactory datasets and tasks with minimal training. The detailed methodology for model fine-tuning and evaluation is described in the subsequent sections.

\subsection{Transfer learning with a foundational model}
As discussed above, high-quality odor datasets are challenging and costly to assemble resulting in limited data availability for building machine olfaction models. Transfer learning addresses this challenge by allowing models trained on larger, related datasets to be adapted for use in data-scarce target domains \cite{weiss2016survey}.

Foundation models in the molecular and chemical sciences provide a powerful framework for transfer learning to improve machine olfaction~\cite{wadell2026foundation}. These models are pretrained on large, diverse chemical datasets and can then be fine-tuned for a wide range of downstream tasks, including those not represented in the original training data \cite{bommasani2021opportunities, awais2023foundational}. In our work, we employ this pretrain-finetune paradigm to leverage molecular representations learned from large-scale pretraining and adapt them for the specific challenges of odor descriptor prediction.

The choice of foundation model is critical to this paradigm. Olfaction is strongly tied to molecular shape, geometry, and stereochemistry, which are only partially captured by SMILES strings or 2D molecular graphs. Uni-Mol2 is particularly suitable for this setting because it is a large-scale 3D molecular foundation model whose architecture explicitly integrates atom-level, graph-level, and geometric information~\cite{ji2024unimol2}. This makes its representation more sensitive to spatial molecular structure. By fine-tuning Uni-Mol2 on the GS-LF dataset, we aim to improve odor descriptor prediction and evaluate its transferability across downstream olfactory tasks.

\subsection{Model fine-tuning and training}

Odor descriptor prediction is formulated as a multi-label classification task, where each molecule may be associated with one or more odor descriptors. A significant challenge in this setting is the pronounced class imbalance within the GS-LF dataset. For example, the most frequent label, ``fruity,'' appears 1,902 times, whereas the rarest label, ``chamomile,'' occurs only 31 times across the dataset's 4,983 molecules. Addressing this imbalance is essential for achieving robust model performance.

To mitigate this, we fine-tuned Uni-Mol2 on the GS-LF training set using a focal loss function, which down-weights the contribution of frequent classes and places greater emphasis on minority classes \cite{ross2017focal}. The focal loss is defined as:
\[
\mathcal{L}_{\text{focal}}(p_t) = -\alpha_t (1 - p_t)^{\gamma} \log(p_t)
\]
where \( p_t \) is the predicted probability of the true class, \( \alpha_t \) is a class balancing factor, and \( \gamma \) is a focusing parameter that reduces the loss contribution of well-classified examples, thereby helping to address class imbalance.

To improve robustness, we ran 50 hyperparameter search trials and constructed an ensemble using the models from the top 10 configurations across five folds, selected based on validation AUROC. Predictions from the 50 models were averaged to obtain the final ensemble predictions. For the first three tasks, the model is used directly for evaluation without additional task-specific training. For mixture prediction task, the molecular embeddings generated by the model are used as inputs for downstream training. We emphasize that downstream training for the mixture task was not deep learning. It only used light-weight classical machine learning.

Our evaluations are based on point estimates computed on the test set and 95\% confidence intervals with lower and upper bounds given by the 2.5th and 97.5th percentiles of the bootstrap estimates. For multi-label classification, we excluded labels with only one class in a bootstrap resample when computing macro AUROC and AUPRC. For macro F1, precision, and recall, we retained all labels and set undefined values to zero.  As positive examples are sparse for some tasks, these rules can shift the bootstrap distribution enough for the percentile interval to fall entirely above or below the point estimate. We discuss cases where the confidence interval excludes the point estimate in the relevant analysis. 5000 bootstrap resamples are used to obtain more stable estimates of the confidence interval bounds.

All experiments were performed on the University of Michigan Great Lakes HPC cluster with a single NVIDIA A40 GPU (48 GB memory). Running 50 Optuna trials with five-fold cross-validation took approximately 120 hours. Ensemble inference on the GS-LF test set took approximately 30 minutes.

\section{Results}

We evaluate our model from two aspects. First, we assess its performance on the core QSOR task of single molecule odor descriptor prediction on the GS-LF test dataset (which was not touched during fine-tuning). Second, we examine the model's transferability across several important downstream olfactory tasks. For comparison, we report the originally published results from POM, a state-of-the-art model that uses graph neural networks on 2D molecular graphs to predict odor descriptors, along with the RF-Mordred baseline reported in the same study, and
our reproduced results using its open-source implementation OpenPOM\cite{sanchez2019machine, lee2023principal, openpom}.  To ensure a fair comparison, we follow POM's evaluation protocol and report results based on a 50-model ensemble, consistent with the original POM setup. To assess the contributions of ensembling and fine-tuning, we also report results for individual fine-tuned Uni-Mol2 models and for a 50-model ensemble with frozen Uni-Mol2 backbones. Notably, our hyperparameter search was substantially smaller in scale than POM's (50 trials versus 500). 

\subsection{Single molecule odor descriptor prediction}
\label{sec:pretrain_transfer_integration}

Table \ref{tab:gslf-task1} presents comprehensive results comparing the models across the original GS-LF dataset. For Uni-Mol2, the 84M-parameter configuration was evaluated, and the best-performing configuration is reported. Increasing the model size to 164M did not improve performance under the same hyperparameter tuning budget (full results are provided in Table S1 of the Supplementary Information). Uni-Mol2 achieves the highest macro AUROC, AUPRC, F1, precision, and recall among the compared models, showing improved performance over both the originally published POM results and the publicly available, fully reproducible OpenPOM baseline. With ensemble size fixed at 50, fine-tuning the Uni-Mol2 backbone increased macro AUROC from 0.8407 to 0.8990 on the test set. The fine-tuned ensemble also outperformed its individual models, which averaged 0.8761 in macro AUROC and 0.3465 in macro AUPRC.

To better assess the performance differences between Uni-Mol2 and OpenPOM, we report paired performance differences across all tasks (full results are provided in Table S2 of Supplementary Information). On GS-LF, Uni-Mol2 outperformed OpenPOM on all five metrics, with all five paired 95\% bootstrap confidence intervals excluding zero.


\begin{table}[ht]
\small
\centering
\caption{Performance comparison on the GS-LF dataset for the multi-label classification task. Each cell reports the point estimate with its 95\% bootstrap confidence interval below. Single-model Uni-Mol2 (FT) results are reported as the mean $\pm$ standard deviation across individual models. For each of 5000 bootstrap resamples of the test molecules, we evaluated all 50 models on the same resampled molecules and averaged their scores for each metric. The POM and RF-Mordred results are reported as published, with intervals based on 1000 bootstrap resamples; we used 5000 bootstrap resamples for all other intervals. In the frozen setting, the pretrained Uni-Mol2 backbone was kept fixed, and only the classification head was trained using focal loss. We applied the same Optuna hyperparameter search and ensembling procedure as in the fine-tuned model to ensure a fair comparison. Best value per metric in bold; ``--'' marks metrics not reported by the original POM study. }
\label{tab:gslf-task1}
\resizebox{\linewidth}{!}{%
\begin{tabular}{lccccc}
\hline
Model & Macro AUROC & Macro AUPRC & Macro F1 & Macro Precision & Macro Recall \\
\hline
POM & 0.8940 & -- & 0.3600 & 0.3790 & 0.3900 \\
              & {\scriptsize [0.8880, 0.9020]} & & {\scriptsize [0.3370, 0.3720]} & {\scriptsize [0.3510, 0.3980]} & {\scriptsize [0.3650, 0.4120]} \\
\addlinespace
RF-Mordred & 0.8500 & -- & 0.3060 & 0.3110 & 0.3930 \\
           & {\scriptsize [0.8380, 0.8600]} & & {\scriptsize [0.2830, 0.3190]} & {\scriptsize [0.2880, 0.3330]} & {\scriptsize [0.3720, 0.4170]} \\
\addlinespace   
OpenPOM & 0.8883 & 0.3748 & 0.3748 & 0.4060 & 0.4049 \\
        & {\scriptsize [0.8801, 0.8963]} & {\scriptsize [0.3720, 0.4122]} & {\scriptsize [0.3485, 0.3893]} & {\scriptsize [0.3748, 0.4257]} & {\scriptsize [0.3803, 0.4297]} \\
\addlinespace
Uni-Mol2 (Frozen) & 0.8407 & 0.2393 & 0.2460 & 0.2267 & 0.4034 \\
                            & {\scriptsize [0.8311, 0.8497]} & {\scriptsize [0.2401, 0.2739]} & {\scriptsize [0.2299, 0.2567]} & {\scriptsize [0.2024, 0.2385]} & {\scriptsize [0.3807, 0.4252]} \\
\addlinespace
Uni-Mol2 (FT) (single) & $0.8761 \pm 0.0022$ & $0.3465 \pm 0.0094$ & $0.3425 \pm 0.0083$ & $0.3627 \pm 0.0158$ & $0.3975 \pm 0.0120$ \\
                      & {\scriptsize [0.8684, 0.8832]} & {\scriptsize [0.3459, 0.3798]} & {\scriptsize [0.3225, 0.3513]} & {\scriptsize [0.3417, 0.3742]} & {\scriptsize [0.3796, 0.4147]} \\
\addlinespace
Uni-Mol2 (FT) & \textbf{0.8990} & \textbf{0.4077} & \textbf{0.3991} & \textbf{0.4288} & \textbf{0.4291} \\
                        & {\scriptsize [0.8916, 0.9060]} & {\scriptsize [0.4015, 0.4434]} & {\scriptsize [0.3720, 0.4122]} & {\scriptsize [0.3962, 0.4476]} & {\scriptsize [0.4056, 0.4520]} \\
\hline
\end{tabular}%
}
\end{table}

\subsubsection{Enantiomer-pair subset analysis}

\begin{figure}[h]
 \centering
 \includegraphics[width=0.5\textwidth]{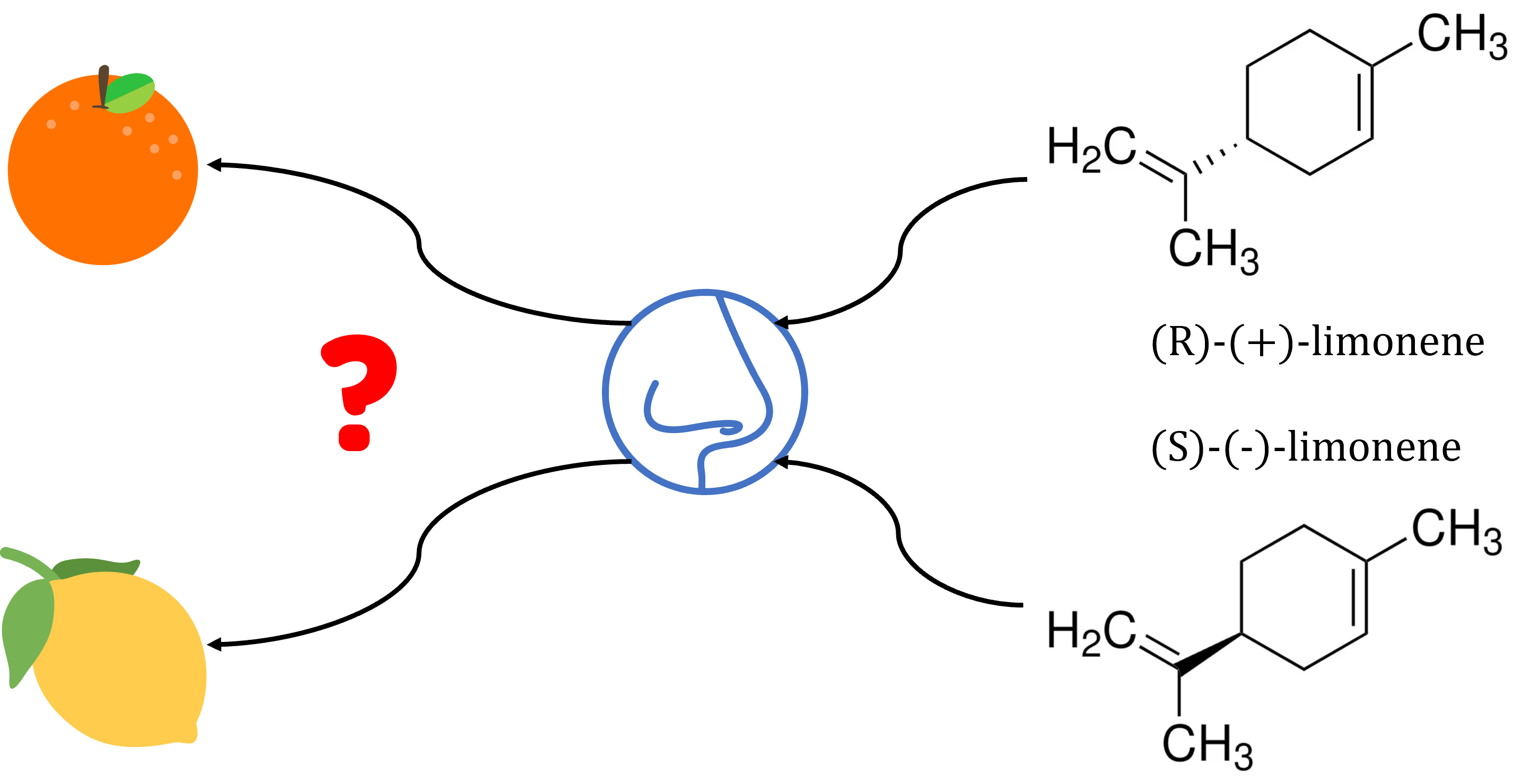}
 \caption{A common misconception regarding the olfactory distinction between limonene enantiomers, highlighting a challenge in machine olfaction.}
 \label{fig:limon}
\end{figure}

As illustrated in Figure~\ref{fig:limon}, the relationship between stereochemistry and olfactory perception remains a fundamental and incompletely understood problem. A classic example is the widespread claim that (S)-(-)-limonene smells primarily of lemon whereas (R)-(+)-limonene smells of orange. However, recent psychophysical studies have shown that this simple dichotomy is not universally valid~\cite{kvittingen2021limonene}. More broadly, enantiomeric pairs span a spectrum of behavior: some exhibit nearly indistinguishable odor profiles, whereas others differ substantially. Consequently, the central computational challenge is not merely recognizing molecular chirality, but predicting when stereochemical differences translate into perceptual differences.

To further evaluate the model's ability to capture stereochemical information, we consider a curated subset of 11 enantiomeric pairs~\cite{king2024transfer}. Both models are evaluated on this subset using the label-wise thresholds learned from GS-LF out-of-fold predictions, without any additional training or threshold tuning. We use RDKit to generate a conformer for each enantiomer independently from its own isomeric SMILES rather than generating one conformer and reflecting the coordinates to obtain its exact mirror image.

The aggregate performance metrics in Table~\ref{tab:enantiomer} show that Uni-Mol2 outperforms OpenPOM on the curated enantiomer subset under the same train-once evaluation protocol used throughout this paper. The clearest gain is in AUPRC, with a paired difference of 0.1187 (95\% CI [0.0574, 0.1853], Table S2). Because all label-wise thresholds are transferred directly from GS-LF without further tuning, these metrics should be interpreted as evidence of transfer rather than as optimized performance on this small benchmark. It is worth noting that the bootstrap intervals for OpenPOM lie above the point estimate for AUPRC and entirely below it for recall. These shifts are partly explained by the exclusion of labels with no positive examples when computing AUPRC and largely explained by the assignment of zero to undefined recall values. Similarly, the bootstrap interval for OpenPOM’s macro AUPRC also lies entirely above its point estimate on the non-overlapping Zhang subset in Table \ref{tab:zhang-task2}. Aggregate metrics cannot distinguish between a model that predicts identical outputs for both enantiomers and one that recognizes stereochemical differences but predicts their perceptual consequences imperfectly. We therefore next introduce a direct measure of stereochemical sensitivity.

\begin{table}[ht]
\small
\centering
\caption{Performance comparison on the enantiomeric test set (11 pairs).  The 95\% confidence intervals (CI) were computed
based on 5000 bootstrap samples of the 22 individual molecules. Threshold-dependent metrics were computed using label-wise thresholds optimized on GS-LF out-of-fold predictions during training, without additional training or threshold tuning on the enantiomeric subset.}
\label{tab:enantiomer}
\begin{tabular*}{\textwidth}{@{\extracolsep{\fill}}lccccc}
\hline
Model & Macro AUROC & Macro AUPRC & Macro F1 &  Macro Precision  & Macro Recall \\
\hline
OpenPOM
& 0.7991 & 0.4082 & 0.0859 & 0.0631 & 0.1568 \\
& {\scriptsize[0.7382, 0.8342]}
& {\scriptsize[0.4185, 0.6378]}
& {\scriptsize[0.0504, 0.0905]}
& {\scriptsize[0.0386, 0.0778]}
& {\scriptsize[0.0889, 0.1505]} \\

\addlinespace

Uni-Mol2 (FT)
& \textbf{0.8264} & \textbf{0.5269} & \textbf{0.0986}
& \textbf{0.0746} & \textbf{0.1616} \\
& {\scriptsize[0.7633, 0.8517]}
& {\scriptsize[0.5264, 0.7033]}
& {\scriptsize[0.0592, 0.1026]}
& {\scriptsize[0.0469, 0.0905]}
& {\scriptsize[0.0935, 0.1505]} \\
\hline
\end{tabular*}
\end{table}

To directly probe stereochemical sensitivity rather than aggregate accuracy, we measure the within-pair divergence of each model's predictions. For every enantiomeric pair, we compute the $L_1$ distance between the predicted descriptor-probability vectors of the two mirror-image molecules. A model that assigns identical predictions to both enantiomers exhibits zero stereochemical sensitivity, whereas larger within-pair divergences indicate that the learned representation recognizes the molecules as distinct.

\begin{figure}[h]
 \centering
 \includegraphics[width=\textwidth]{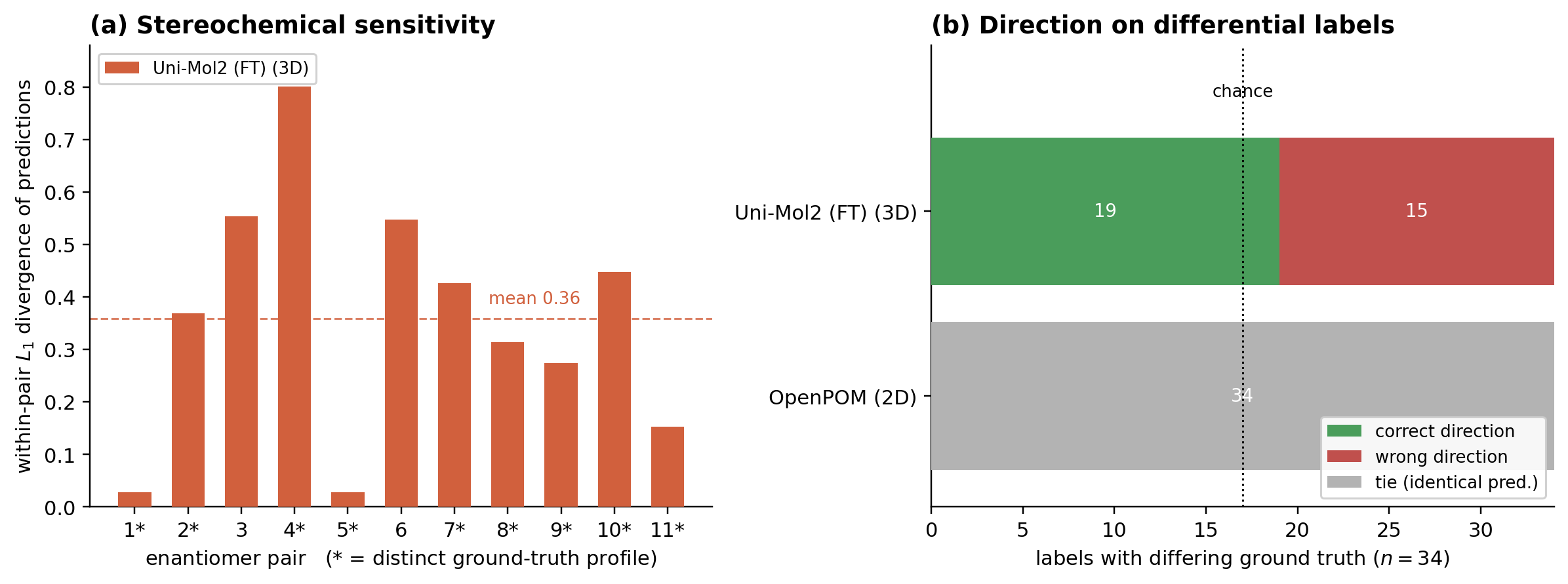}
 \caption{Stereochemical discrimination on the 11 enantiomeric pairs. (a) Per-pair within-pair $L_1$ divergence between Uni-Mol2's predicted probability vectors for the two mirror-image enantiomers (bars; dashed line, mean $=0.36$); asterisks mark pairs whose ground-truth profiles differ. OpenPOM (2D) assigns identical predictions to both enantiomers of every pair (within-pair $L_1 = 0$) and is therefore not plotted. (b) On the labels whose ground truth differs between the two enantiomers ($n=34$), Uni-Mol2 ranks the truly-positive enantiomer higher for 19 (vs.\ 15 wrong; the dotted line marks 17 of 34 labels (50\%)), whereas OpenPOM produced ties on all 34 labels.}
 \label{fig:enantiomer-discrim}
\end{figure}

OpenPOM assigned identical predictions to both members of all 11 enantiomer pairs (within-pair $L_1 = 0$), including ties on all 34 labels that differed within a pair. In contrast, Uni-Mol2 produces distinct predictions for every enantiomeric pair (mean within-pair $L_1 = 0.36$; paired Wilcoxon $p<0.001$), demonstrating that the learned molecular representation is sensitive to stereochemistry (Figure~\ref{fig:enantiomer-discrim} (a)). Uni-Mol2 takes both 3D coordinates and explicit stereochemical features as input, and the relative contributions of each to this sensitivity remain an open question.

Representation sensitivity alone, however, does not imply correct prediction of enantiomer-specific odor perception. We therefore ask a second question: when the ground-truth odor descriptors differ between two enantiomers, does the model assign the higher probability to the correct molecule? Uni-Mol2 assigned a higher probability to the correct enantiomer for 19 of the 34 differential labels (55.9\%, Figure~\ref{fig:enantiomer-discrim} (b)). The 95\% percentile confidence interval was 44.8\%, 69.0\%, based on 5000 bootstrap resamples of whole enantiomer pairs. Resampling whole pairs kept all labels from the same pair together.

\begin{figure}[h]
 \centering
 \includegraphics[width=\textwidth]{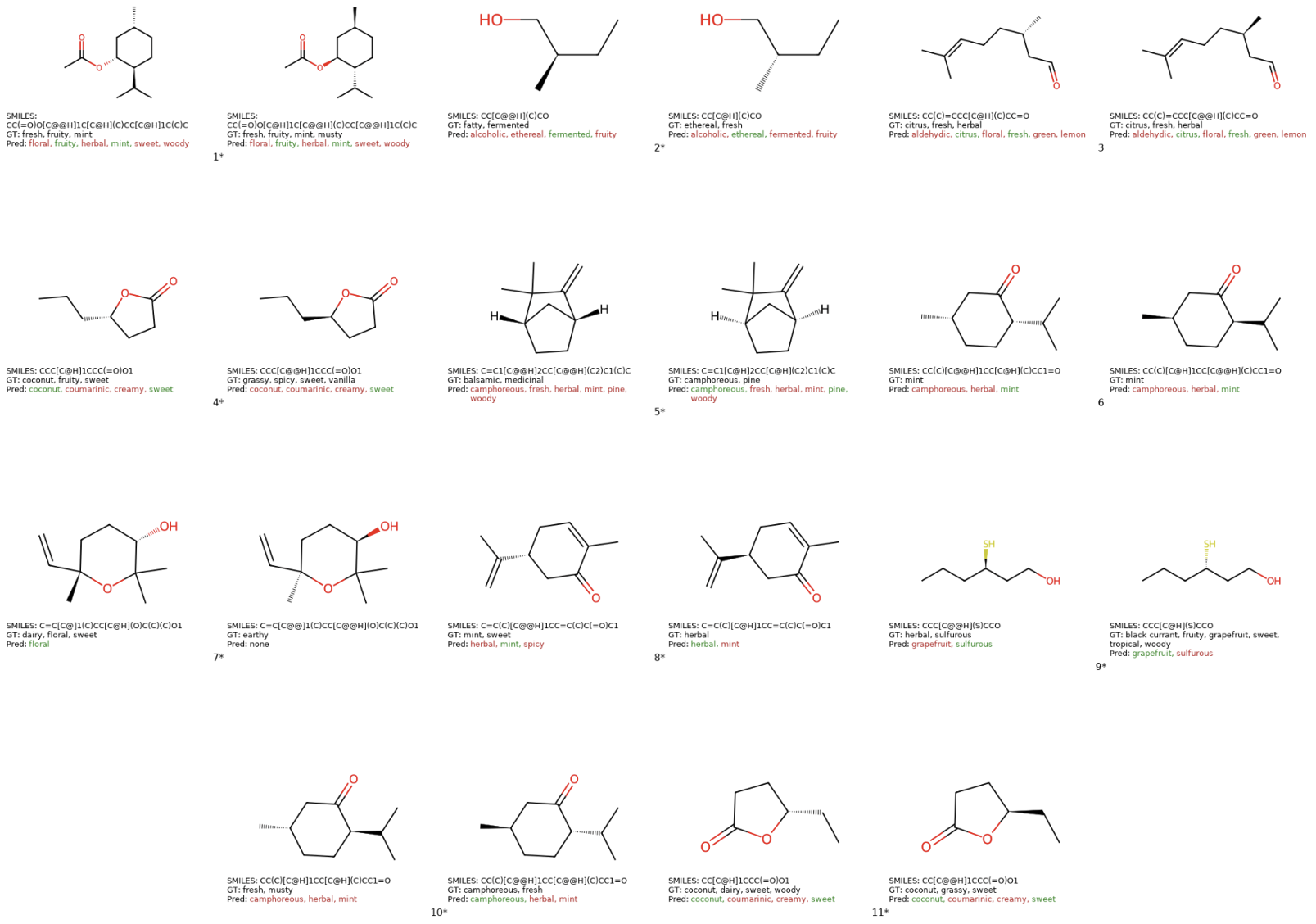}
 \caption{Eleven enantiomeric pairs with distinct or similar sensory profiles (* means distinct ground-truth profile). Green labels indicate correct predictions, while red labels denote incorrect predictions made by the Uni-Mol2 model trained on the GS-LF dataset.}
 \label{fig:enantiomers}
\end{figure}

Figure~\ref{fig:enantiomers} presents the Uni-Mol2 predictions for the enantiomeric pairs. Consistent with the aggregate analysis above, Uni-Mol2 typically predicts highly similar odor profiles for the two mirror-image molecules, even when their annotated odor descriptors differ substantially. For example, pair no.\ 4 receives identical predicted profiles (``coconut, coumarinic, creamy, sweet'') despite differing ground-truth annotations, and pair no.\ 10 likewise receives nearly identical predictions (``camphoreous, herbal, mint'') despite different annotated odor profiles. These qualitative examples illustrate why stereochemical sensitivity alone is insufficient for accurate enantiomer-specific odor prediction.

\subsection{Cross-dataset descriptor prediction}

Odor descriptor datasets are limited in scale and often assembled from different sources. Even when datasets share similar descriptor vocabularies, differences in molecule selection, label coverage, and annotation collection can affect model performance. Therefore, performance on GS-LF alone does not fully establish the model's generalizability.

To evaluate whether the model generalizes beyond its original training distribution, we further conduct a cross-dataset evaluation on the test set from Zhang et al. dataset~\cite{zhang2024deep}. 
Derived from the Pyrfume repository~\cite{hamel2024pyrfume}, the dataset includes annotations for 118 odor descriptors, 108 of which overlap with those in GS-LF. We apply the models to the test set, which contains 837 molecules, to predict the 108 shared descriptors. Because some molecules in the Zhang test set also appear in the GS-LF training set, we report results on both the complete test set and a non-overlapping subset obtained by removing these molecules.

The results in Table~\ref{tab:zhang-task2} show that Uni-Mol2 achieves a high AUROC of 0.8975 on the non-overlap test set, suggesting strong transferability to external odor descriptor datasets. Compared to OpenPOM, our model consistently performs better on both the complete Zhang test set and the non-overlapping set, which highlights the model's ability to capture transferable structural and chemical features through transfer learning. On both the complete Zhang test set and the non-overlapping subset, the paired 95\% confidence intervals for gains in AUPRC, F1, and recall were all above zero (Table S2). We also examined how performance varied with structural similarity to the training set using fingerprint-based Tanimoto distance (full results are provided in Table S3). As the minimum chemical distance increases, AUROC declines for both models, and Uni-Mol2’s AUPRC advantage over OpenPOM narrows to 0.0028 in the most distant subset.


\begin{table}[ht]
\small
\centering
\caption{Performance comparison on the Zhang test set with and without overlapping molecules. The 95\% confidence intervals (CI) were computed based on 5000 bootstrap samples. Thresholds were optimized on GS-LF out-of-fold predictions and applied directly to the Zhang test set.}
\label{tab:zhang-task2}

\resizebox{\textwidth}{!}{%
\begin{tabular}{llccccc}
\hline
Model & Setting & Macro AUROC & Macro AUPRC & Macro F1 & Macro Precision & Macro Recall \\
\hline
OpenPOM & non-overlap
& 0.8937 & 0.2937 & 0.2453 & 0.2075 & 0.4080 \\
&
& {\scriptsize[0.8725, 0.9106]}
& {\scriptsize[0.2995, 0.3741]}
& {\scriptsize[0.2042, 0.2593]}
& {\scriptsize[0.1798, 0.2305]}
& {\scriptsize[0.3339, 0.4256]} \\

\addlinespace

Uni-Mol2 (FT) & non-overlap
& \textbf{0.8975} & \textbf{0.3609} & \textbf{0.2696}
& \textbf{0.2264} & \textbf{0.4510} \\
&
& {\scriptsize[0.8727, 0.9138]}
& {\scriptsize[0.3467, 0.4306]}
& {\scriptsize[0.2214, 0.2864]}
& {\scriptsize[0.1892, 0.2506]}
& {\scriptsize[0.3630, 0.4678]} \\

\addlinespace

OpenPOM & complete
& 0.9227 & 0.3438 & 0.3357 & 0.2740 & 0.5331 \\
&
& {\scriptsize[0.9096, 0.9366]}
& {\scriptsize[0.3425, 0.4046]}
& {\scriptsize[0.3018, 0.3513]}
& {\scriptsize[0.2500, 0.2973]}
& {\scriptsize[0.4898, 0.5655]} \\

\addlinespace

Uni-Mol2 (FT) & complete
& \textbf{0.9291} & \textbf{0.3846} & \textbf{0.3621}
& \textbf{0.2949} & \textbf{0.5796} \\
&
& {\scriptsize[0.9153, 0.9428]}
& {\scriptsize[0.3824, 0.4469]}
& {\scriptsize[0.3283, 0.3788]}
& {\scriptsize[0.2712, 0.3180]}
& {\scriptsize[0.5344, 0.6124]} \\
\hline
\end{tabular}%
}
\end{table}

\subsection{Odor detection: predict if a molecule is odorous}

Odorous-versus-odorless prediction is a canonical binary olfactory prediction task. Mayhew et al.~\cite{mayhew2022transport} showed that this task can be accurately modeled using transport-related molecular properties and released a curated benchmark that has become widely used for evaluating odor detection, i.e., predicting whether a given molecule is odorous or odorless to humans. We therefore use the Mayhew et al. dataset, which comprises 1,924 molecules, including 1,615 odorous and 309 odorless compounds. After excluding molecules present in the GS-LF training set based on canonical SMILES, the non-overlapping subset comprised 1,378 of the 1,924 test molecules. We additionally report performance on the reduced non-overlapping test set to eliminate any potential data leakage.

As shown in Table~\ref{tab:Odor}, Uni-Mol2 consistently achieved higher AUROC and substantially improved recall and F1 score relative to OpenPOM on both the complete and non-overlapping test sets. Despite its higher F1 and recall, Uni-Mol2 had lower AUPRC than OpenPOM for the odorless class on both the complete test set and the non-overlapping subset. The paired analysis supports gains in F1 and recall on both test sets, while the interval for the AUPRC difference is entirely below zero only on the complete test set (Table S2). For the structural similarity analysis, AUROC declines for both models as the distance threshold increases, while OpenPOM has higher AUPRC than Uni-Mol2 in every subset (Table S3). Overall, the improved balanced classification performance (F1) and higher detection rate for odorless molecules (recall) indicate that the learned molecular representation transfers effectively to binary odor detection.

\begin{table}[ht]
\small
\centering
\caption{Performance comparison on held-out odorous/odorless test set with and without overlapping molecules. The non-overlap subset excludes molecules present in the GS-LF training set based on canonical SMILES. The 95\% confidence intervals (CI) were computed based on 5000 bootstrap resamples. Thresholds were optimized on GS-LF out-of-fold predictions and applied directly to the test set. Metrics are computed for the positive class (odorless). The odorless score is the predicted probability from the dedicated odorless output among the 138 GS-LF labels.}
\label{tab:Odor}

\resizebox{\textwidth}{!}{%
\begin{tabular}{llccccc}
\hline
Model & Setting & AUROC & AUPRC & F1 & Precision & Recall \\
\hline

OpenPOM & non-overlap
& 0.7727 & \textbf{0.4423} & 0.3108
& \textbf{0.5041} & 0.2246 \\
& &
{\scriptsize[0.7429, 0.7994]} &
{\scriptsize[0.3858, 0.5013]} &
{\scriptsize[0.2513, 0.3671]} &
{\scriptsize[0.4160, 0.5905]} &
{\scriptsize[0.1765, 0.2731]} \\

\addlinespace
Uni-Mol2 (FT) & non-overlap
& \textbf{0.7897} & 0.4028 & \textbf{0.4328}
& 0.4462 & \textbf{0.4203} \\
& &
{\scriptsize[0.7636, 0.8142]} &
{\scriptsize[0.3533, 0.4620]} &
{\scriptsize[0.3760, 0.4838]} &
{\scriptsize[0.3825, 0.5062]} &
{\scriptsize[0.3626, 0.4775]} \\

\addlinespace
OpenPOM & complete
& 0.8374 & \textbf{0.4669} & 0.3251 & \textbf{0.5373} & 0.2330 \\
& &
{\scriptsize[0.8152, 0.8586]} &
{\scriptsize[0.4129, 0.5255]} &
{\scriptsize[0.2673, 0.3814]} &
{\scriptsize[0.4552, 0.6187]} &
{\scriptsize[0.1866, 0.2828]} \\

\addlinespace
Uni-Mol2 (FT) & complete
& \textbf{0.8507} & 0.4234 & \textbf{0.4444} & 0.4710 & \textbf{0.4207} \\
& &
{\scriptsize[0.8312, 0.8690]} &
{\scriptsize[0.3775, 0.4803]} &
{\scriptsize[0.3927, 0.4942]} &
{\scriptsize[0.4109, 0.5308]} &
{\scriptsize[0.3661, 0.4778]} \\

\hline
\end{tabular}%
}
\end{table}

\subsection{Mixture discrimination}

To extend our evaluation beyond single-molecule odorants, we considered odor mixtures. Naturally occurring odors are almost always mixtures, but modeling them is challenging because the percept of a mixture is generally not a simple combination of the percepts of its constituent molecules. We focus on the task of predicting the normalized perceptual discriminability between two odor mixtures, represented by a score in the interval $[0,1]$, where larger values indicate that the two mixtures are more perceptually distinguishable (i.e., less similar).

Previous approaches to this task can be broadly divided into two categories:

\begin{enumerate}
\item \textbf{Direct methods}: These methods predict perceptual similarity directly from chemically derived descriptors of odor mixtures (e.g., Dragon descriptors), without constructing an intermediate perceptual representation~\cite{snitz2013predicting,ravia2020measure}. Although conceptually simple, they have been shown to underperform methods based on learned intermediate representations~\cite{dhurandhar2023expansive}.

\item \textbf{Representation-based methods}: These methods first construct an intermediate representation for each molecule or mixture and then learn a metric relating differences between these representations to human perceptual similarity. Dhurandhar \textit{et al.}~\cite{dhurandhar2023expansive}, for example, predict semantic odor descriptors for individual molecules, aggregate them to obtain mixture-level representations, and perform metric learning on the resulting semantic features. Tom \textit{et al.}~\cite{tom2025pommix} introduced POMMix, which uses attention to combine molecular embeddings into a mixture representation and predicts perceptual distance using a cosine-based prediction head.
\end{enumerate}

Our approach follows the general representation-learning framework of Dhurandhar \textit{et al.}~\cite{dhurandhar2023expansive}. We represent each molecule by the fine-tuned Uni-Mol2 model’s predicted probabilities for the 138 GS-LF odor descriptors and average these vectors across the constituent molecules of each mixture. This averaging gives each component equal weight regardless of concentration and does not explicitly encode the number of components. We use the elementwise squared differences between the two mixture vectors as inputs to a regressor that predicts perceptual discriminability.

To fully utilize the limited mixture data, we performed five-fold cross-validation  within each dataset at the mixture-pair level, and pooled training pairs across all four datasets in each round. Under this scheme, individual mixtures appear in both training and test pairs. We evaluated multiple regression models, including ElasticNet, random forest, and gradient boosting, and reported performance separately on the held-out pairs from each dataset as well as on the combined held-out predictions across all datasets. This setup evaluates whether molecular representations learned from single-molecule odor descriptor prediction transfer to predicting perceptual discriminability on held-out mixture pairs.

The comparative results, including both RMSE and Pearson correlation, are summarized in Table~\ref{tab:mixture_eval}. On the combined dataset, Uni-Mol2 with GBT achieved the lowest RMSE and highest Pearson correlation among the evaluated model–regressor combinations. In comparisons using the same regressor, paired bootstrap intervals supported improvements in both metrics with ElasticNet and a reduction in RMSE with GBT (Table S2, Panel B). Uni-Mol2 also achieves the best performance on three of the four individual datasets (Snitz~1, Snitz~2, and Ravia), demonstrating that representations learned from single-molecule odor descriptor prediction transfer effectively to the distinct problem of odor mixture discriminability. 

The only exception is the Bushdid dataset, where OpenPOM performs slightly better. Bushdid differs from the other benchmarks in its experimental design: the target is the fraction of subjects correctly identifying the odd mixture in a triangle discrimination test rather than a similarity rating converted to discriminability. The reason for this difference remains an interesting direction for future investigation. 

To assess generalization to unseen datasets, we also report leave-one-dataset-out results in Table S4 of the Supplementary Information. When each dataset is held out in turn and the regressors are trained on the other three, Uni-Mol2 with GBT achieves lower RMSE and higher Pearson correlation than OpenPOM with GBT on both Bushdid and Snitz 1. 

Overall, these results demonstrate that a molecular representation learned from a single canonical olfactory task generalizes not only across datasets but also to the qualitatively different task of predicting perceptual discriminability between odor mixtures.

\begin{table}[ht]
\centering
\caption{Performance comparison on odor mixture datasets. Lower RMSE and
higher Pearson correlation indicate better prediction accuracy. Combined
results are computed from the concatenated out-of-fold predictions across
all four datasets. The 95\% confidence intervals (CI) were computed based on 5000 bootstrap resamples.}
\label{tab:mixture_eval}

\newcommand{\mixci}[2]{{\scriptsize[#1,\,#2]}}
\resizebox{\textwidth}{!}{%
\begin{tabular}{ll cc cc cc cc cc}
\toprule
\multirow{2}{*}{Model} & \multirow{2}{*}{Regressor}
& \multicolumn{2}{c}{Combined}
& \multicolumn{2}{c}{Bushdid}
& \multicolumn{2}{c}{Snitz 1}
& \multicolumn{2}{c}{Snitz 2}
& \multicolumn{2}{c}{Ravia} \\
\cmidrule(lr){3-4}
\cmidrule(lr){5-6}
\cmidrule(lr){7-8}
\cmidrule(lr){9-10}
\cmidrule(lr){11-12}
& & RMSE & Pearson & RMSE & Pearson & RMSE & Pearson
  & RMSE & Pearson & RMSE & Pearson \\
\midrule

\multirow{6}{*}{OpenPOM}
& ElasticNet
& 0.148 & 0.343 & 0.172 & 0.348 & 0.115 & 0.478
& 0.122 & 0.353 & 0.141 & 0.362 \\
& &
\mixci{0.140}{0.155} & \mixci{0.288}{0.398} &
\mixci{0.160}{0.184} & \mixci{0.238}{0.452} &
\mixci{0.105}{0.125} & \mixci{0.381}{0.571} &
\mixci{0.108}{0.136} & \mixci{0.164}{0.528} &
\mixci{0.114}{0.167} & \mixci{0.115}{0.594} \\

& RF
& 0.131 & 0.557 & \textbf{0.145} & \textbf{0.581} & 0.109 & 0.555
& 0.120 & 0.605 & 0.132 & 0.411 \\
& &
\mixci{0.124}{0.138} & \mixci{0.499}{0.611} &
\mixci{0.134}{0.156} & \mixci{0.503}{0.655} &
\mixci{0.098}{0.119} & \mixci{0.462}{0.645} &
\mixci{0.104}{0.136} & \mixci{0.453}{0.733} &
\mixci{0.104}{0.160} & \mixci{0.177}{0.622} \\

& GBT
& 0.132 & 0.542 & 0.147 & 0.556 & 0.109 & 0.519
& 0.115 & 0.516 & 0.138 & 0.304 \\
& &
\mixci{0.125}{0.140} & \mixci{0.482}{0.600} &
\mixci{0.135}{0.160} & \mixci{0.467}{0.640} &
\mixci{0.098}{0.120} & \mixci{0.419}{0.614} &
\mixci{0.099}{0.131} & \mixci{0.366}{0.648} &
\mixci{0.108}{0.167} & \mixci{0.052}{0.543} \\
\midrule

\multirow{6}{*}{Uni-Mol2 (FT)}
& ElasticNet
& 0.144 & 0.396 & 0.169 & 0.379 & 0.113 & 0.494
& 0.112 & 0.568 & 0.138 & 0.385 \\
& &
\mixci{0.137}{0.152} & \mixci{0.342}{0.451} &
\mixci{0.157}{0.181} & \mixci{0.267}{0.486} &
\mixci{0.103}{0.123} & \mixci{0.398}{0.595} &
\mixci{0.099}{0.124} & \mixci{0.436}{0.683} &
\mixci{0.111}{0.166} & \mixci{0.163}{0.603} \\

& RF
& 0.129 & 0.573 & 0.147 & 0.566 & 0.101 & 0.648
& 0.114 & \textbf{0.674} & \textbf{0.131} & \textbf{0.424} \\
& &
\mixci{0.122}{0.137} & \mixci{0.517}{0.628} &
\mixci{0.135}{0.158} & \mixci{0.481}{0.645} &
\mixci{0.092}{0.111} & \mixci{0.569}{0.721} &
\mixci{0.099}{0.128} & \mixci{0.565}{0.771} &
\mixci{0.102}{0.160} & \mixci{0.168}{0.650} \\

& GBT
& \textbf{0.128} & \textbf{0.578} & 0.149 & 0.538
& \textbf{0.095} & \textbf{0.673} & \textbf{0.105} & 0.656
& 0.132 & 0.416 \\
& &
\mixci{0.120}{0.135} & \mixci{0.523}{0.633} &
\mixci{0.137}{0.161} & \mixci{0.449}{0.622} &
\mixci{0.086}{0.104} & \mixci{0.597}{0.739} &
\mixci{0.091}{0.119} & \mixci{0.548}{0.746} &
\mixci{0.105}{0.158} & \mixci{0.203}{0.612} \\
\bottomrule
\end{tabular}%
}
\end{table}

\section{Toward foundation models for both molecules and odor descriptors}

We conclude by analyzing common error patterns in GS-LF prediction to motivate a future direction in which foundation models are used to learn transferable representations for both molecules and odor descriptors. 
Rather than treating odor descriptors as independent binary labels, future models could learn joint representation spaces for molecules and odor descriptors, enabling prediction through a learned compatibility function. This approach can potentially improve prediction accuracy while enabling zero-shot generalization to previously unseen odor descriptors.

We define a confusing label pair $(\ell_{\mathrm{true}}, \ell_{\mathrm{pred}})$ as an error in which a molecule annotated with descriptor $\ell_{\mathrm{true}}$ is instead assigned descriptor $\ell_{\mathrm{pred}}$. Operationally, this corresponds to a false negative for $\ell_{\mathrm{true}}$ occurring simultaneously with a false positive for $\ell_{\mathrm{pred}}$ on the same molecule. Table~\ref{tab:label-prediction} lists the ten most frequently occurring confusing label pairs in the GS-LF test set (996 molecules).

\begin{table}[ht]
\small
\centering
\caption{Top 10 most frequently confused label pairs in the GS-LF test set and their co-occurence counts.}
\label{tab:label-prediction}

\begin{tabular*}{\textwidth}{@{\extracolsep{\fill}}llcc}
    \hline
    True Label & Predicted Label & Molecule Count & Co-occurrence Count \\
    \hline
    Green & Sweet & 24 & 340 \\
    Spicy & Sweet & 23 & 162\\
    Fruity & Fresh & 23 & 200\\
    Herbal & Sweet & 21 & 209\\
    Fruity & Floral & 19 & 396 \\
    Fruity & Herbal & 19 & 258 \\
    Oily & Sweet & 19 & 93 \\
    Fruity & Sweet & 18 & 596 \\
    Green & Floral & 16 & 317\\
    Rose & Sweet & 16 & 117\\
    \hline
\end{tabular*}
\end{table}

\begin{figure}[h]
\centering
  \includegraphics[width=0.7\textwidth]{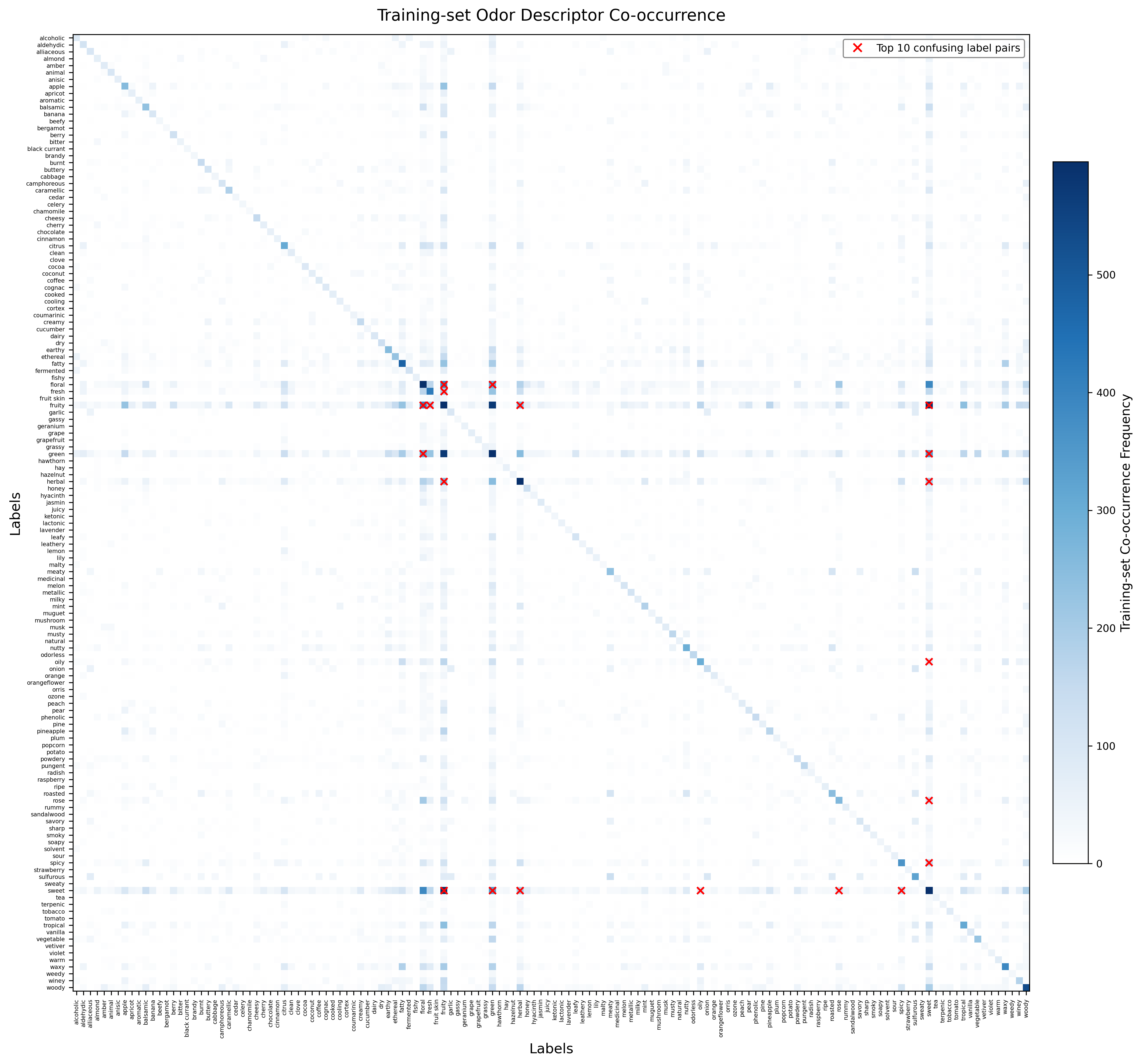}
  \caption{Co-occurrence matrix for labels. The top 10 confusing label pairs are indicated by red crosses. Diagonal values represent self-co-occurrence frequencies.}
  \label{fig:co_occurrence}
\end{figure}

We found that the most frequently confused label pairs exhibit notably high co-occurrence in the GS-LF training set (Figure~\ref{fig:co_occurrence}). The most confused pair, \textit{green} and \textit{sweet}, co-occurs 340 times. Even the least co-occurring pair among these top confusions, \textit{oily} and \textit{sweet}, co-occurs 93 times. On average, the top 10 confusing pairs have a co-occurrence value of 268.8, substantially higher than the dataset’s average pairwise co-occurrence of 5.74. Across all 9,453 label pairs, pairwise confusion frequency showed a positive correlation with label co-occurrence, with a Spearman correlation of 0.596. These results suggest that prediction errors exhibit clear statistical structure: frequent model confusions partly reflect the intrinsic overlap among odor descriptors. 

\begin{figure}[h]
\centering
  \includegraphics[width=0.8\textwidth]{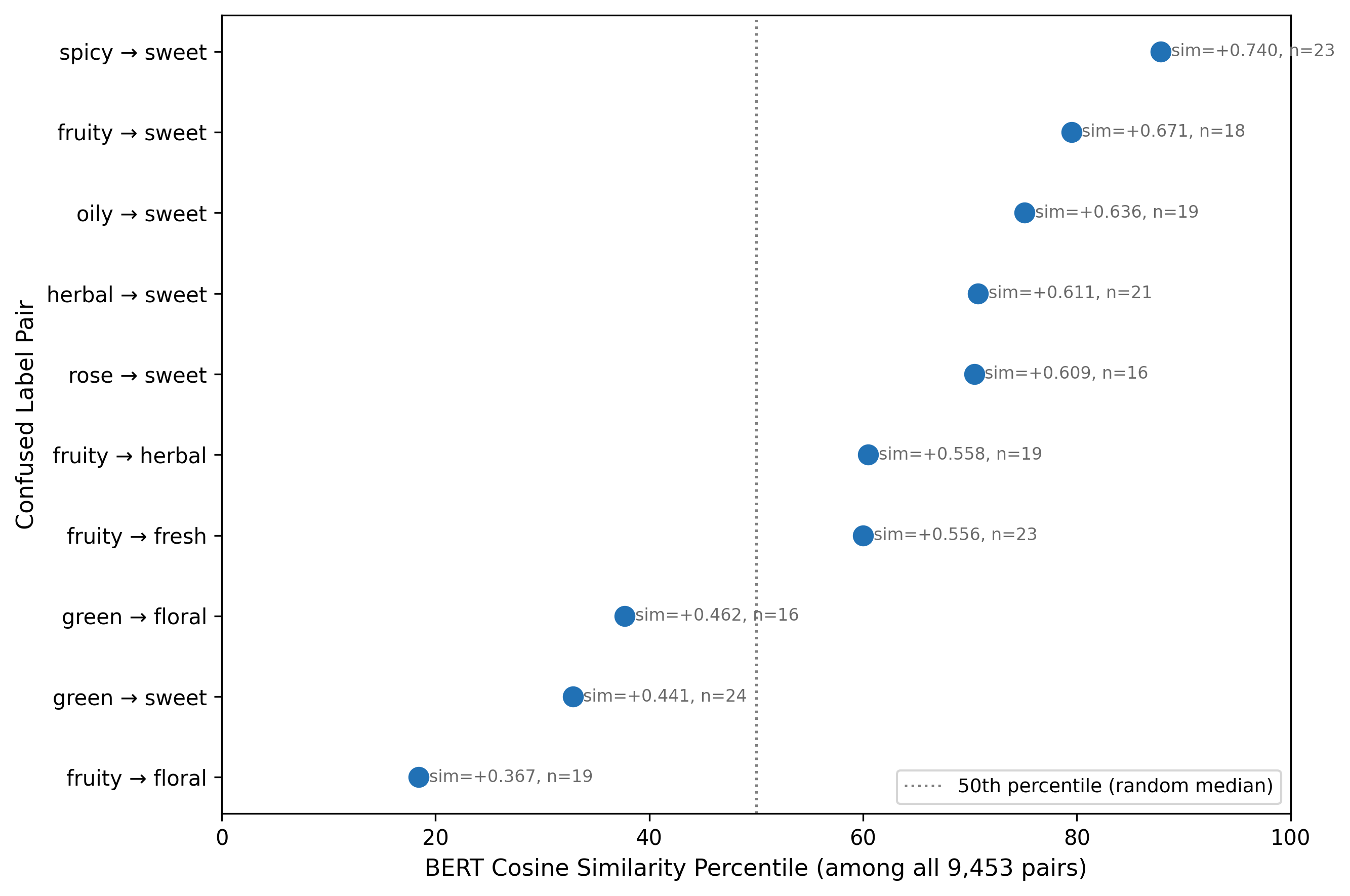}
  \caption{BERT cosine similarity percentiles of the top 10 confusing label pairs. Label pairs are ordered by similarity percentile, and the gray dotted line indicates the 50th percentile of all 9,453 label pairs.}
  \label{fig:bert}
\end{figure}

We next investigated whether this statistical structure could be explained by semantic similarity among odor descriptors. To examine whether the most frequent confusion pairs were close in a language embedding space, we generated BERT embeddings for all descriptors and ranked the cosine similarity of each of the top 10 confusion pairs relative to all 9,453 unique label pairs (Figure~\ref{fig:bert}). The most frequent confusion pairs had a mean similarity percentile of approximately 59, slightly above the chance-level expectation of 50. However, this deviation was not statistically significant in a permutation test ($p=0.16$). Thus, semantic similarity as measured by BERT provided only weak evidence for explaining the observed confusion structure.

The observed pattern of prediction errors suggests that odor descriptors should not be viewed as independent categorical labels. Instead, they exhibit both statistical structure, through frequent co-occurrence, and possible semantic structure, through their proximity in a language embedding space, although the evidence for the latter is weak. While the present work focuses on learning transferable representations of molecules using a molecular foundation model, these observations point to a complementary direction: learning transferable representations of odor descriptors. Future models could jointly embed molecules and odor descriptors and learn a compatibility function between the two representation spaces, replacing multilabel classification with a fixed label space with prediction based on compatibility between molecular and descriptor representations. Such a formulation would naturally capture relationships among descriptors, enable sharing of statistical strength across related labels, and potentially support zero-shot prediction of previously unseen odor descriptors.

\section{Discussion}

Our results support the conclusion that a molecular foundation model fine-tuned on a single canonical olfactory task can yield representations that transfer across several distinct machine olfaction settings. On the GS-LF benchmark, Uni-Mol2 improves over both the published POM results and our reproduced OpenPOM baseline across the reported single-molecule prediction metrics. The same fine-tuned representation also transfers to an external odor-descriptor dataset, to binary odorous-versus-odorless classification, and to perceptual prediction for odor mixtures, without additional deep-learning fine-tuning on these downstream tasks.

These evaluations probe complementary forms of generalization. The Zhang dataset tests cross-dataset transfer under a partially overlapping descriptor vocabulary, while the Mayhew et al.\ benchmark asks whether the representation supports the odorous vs.\ odorless binary prediction problem under substantial class imbalance. The mixture evaluation goes further by moving beyond single-molecule prediction to perceptual discriminability between multicomponent odors. Uni-Mol2 performs best on the combined mixture benchmark and on three of the four individual mixture datasets, indicating that the representation learned from GS-LF captures information that remains useful after both the input structure and prediction target have changed. The Bushdid dataset is the one exception, where OpenPOM performs slightly better. Because Bushdid uses a triangle-discrimination paradigm, whereas the Snitz datasets originate from explicit similarity ratings and Ravia uses a different discrimination protocol, the source of this discrepancy remains unclear and merits further investigation.

The enantiomer analysis provides a more direct view of how the two models respond to stereochemical differences. Aggregate metrics place Uni-Mol2 ahead of OpenPOM on the curated enantiomer subset, but the within-pair analysis reveals a sharper distinction. OpenPOM assigns identical predictions to mirror-image molecules. Uni-Mol2, by contrast, produces distinct predictions for every enantiomeric pair. Uni-Mol2 ranked the correct enantiomer higher for 19 of 34 differential labels, a preliminary result given the dependence among labels within each pair. A comparison with a 2D graph model that explicitly encodes chirality would help isolate the contribution of 3D geometry to enantiomer-specific odor prediction. A complementary direction for future work would be to examine whether Uni-Mol2 produces different predictions for a molecule and its mirror-image conformer.

Taken together, the results suggest that chemically pretrained molecular representations capture transferable information relevant to multiple aspects of olfactory perception, rather than merely specializing to the GS-LF benchmark. At the same time, the modest size of many improvements and the remaining failures on enantiomer-specific perception and Bushdid mixture discrimination indicate that molecular representation quality is only one part of the problem. Future work should investigate which properties of molecular foundation models---including three-dimensional pretraining, scale, corpus composition, and pretraining objective---are most important for transfer to olfaction.

A second limitation lies on the output side of the prediction problem. The present models treat odor descriptors as independent categorical labels, despite their substantial co-occurrence and potential semantic structure. The structured confusion patterns observed on GS-LF motivate future models that learn representations for both molecules and odor descriptors and predict through a compatibility function between the two spaces. Such a formulation could share statistical strength across related descriptors and, importantly, enable zero-shot prediction for odor descriptors that were not present in the GS-LF training vocabulary.

\section{Conclusion}

We have shown that a molecular foundation model fine-tuned on a single canonical olfactory prediction task learns representations that transfer across multiple machine olfaction problems, including cross-dataset descriptor prediction, odorous vs.\ odorless classification, odor mixture discriminability, and stereochemical evaluation. These results suggest that modern molecular foundation models provide a strong foundation for transferable olfactory prediction.

Beyond the empirical improvements reported here, our experiments reveal two broader opportunities for the field. First, understanding what properties of molecular foundation models give rise to transferable olfactory representations remains an open scientific question. Second, our analysis of descriptor confusions suggests that future machine olfaction systems should learn representations for odor descriptors in addition to molecules, replacing closed-vocabulary multilabel prediction with compatibility-based prediction between molecular and descriptor representations.

\section*{Author contributions}

\textbf{Yikun Han:} Data curation, Formal analysis, Investigation, Methodology, Software, Validation, Visualization, Writing – original draft. \textbf{Yi Wang:} Formal analysis, Methodology, Software, Validation, Visualization, Writing –  review \& editing. \textbf{Neil Mankodi:} Methodology, Software. \textbf{Stephen Yang:} Formal analysis, Software, Validation. \textbf{Ambuj Tewari:} Conceptualization, Funding acquisition, Methodology, Project administration, Resources, Supervision, Validation, Writing – review \& editing.

\section*{Conflicts of interest}
There are no conflicts to declare.

\section*{Data availability}
The code developed and datasets used in this paper are publicly available in a GitHub repository at \href{https://github.com/ywee-25/A-General-Purpose-Molecular-Foundation-Model-Transfers-Across-Diverse-Olfactory-Tasks}{this URL}. Supplementary information (SI) is available.

\section*{Acknowledgements}

We gratefully acknowledge the support of two New Initiatives/New Instruction (NINI) Grants from LSA Technology Services at the University of Michigan, awarded in 2023 and 2025. We also thank the Uni-Mol team for their prompt and insightful responses to our questions on the Uni-Mol GitHub repository, which greatly facilitated this research. The majority of this work was carried out while Yikun Han, Neil Mankodi, and Stephen Yang were students at the University of Michigan.

\bibliographystyle{unsrtnat}
\bibliography{rsc}

\end{document}



\pagestyle{fancy}
\thispagestyle{plain}

\fancypagestyle{plain}{
    \renewcommand{\headrulewidth}{0pt}
}

\makeFNbottom

\makeatletter
\renewcommand\LARGE{\@setfontsize\LARGE{15pt}{17}}
\renewcommand\Large{\@setfontsize\Large{12pt}{14}}
\renewcommand\large{\@setfontsize\large{10pt}{12}}
\renewcommand\footnotesize{\@setfontsize\footnotesize{7pt}{10}}
\makeatother

\renewcommand{\thefootnote}{\fnsymbol{footnote}}

\renewcommand\footnoterule{%
    \vspace*{1pt}%
    \color{cream}%
    \hrule width 3.5in height 0.4pt%
    \color{black}%
    \vspace*{5pt}%
}

\makeatletter
\renewcommand\@makefntext[1]{%
    \noindent
    \makebox[0pt][r]{\@thefnmark\,}#1
}
\makeatother

\sectionfont{\sffamily\Large}
\subsectionfont{\normalsize}
\subsubsectionfont{\bfseries}

\setstretch{1.125}
\setlength{\skip\footins}{0.8cm}
\setlength{\footnotesep}{0.25cm}
\setlength{\jot}{10pt}

\titlespacing*{\section}{0pt}{4pt}{4pt}
\titlespacing*{\subsection}{0pt}{15pt}{1pt}

\fancyhead{}
\fancyfoot{}
\fancyfoot[C]{\footnotesize\sffamily\thepage}

\renewcommand{\headrulewidth}{0pt}
\renewcommand{\footrulewidth}{0pt}

\setlength{\arrayrulewidth}{1pt}

\setcounter{table}{0}
\renewcommand{\thetable}{S\arabic{table}}


\sffamily

\begin{center}

{\LARGE\textbf{Supplementary Information}}

\vspace{0.8cm}

{\Large\textbf{
A General-Purpose Molecular Foundation Model Transfers Across Diverse
Olfactory Tasks
}}

\vspace{0.5cm}

{\large
Yikun Han,\textit{$^{a,\ddag}$}
Yi Wang,\textit{$^{a,\ddag}$}
Neil Mankodi,\textit{$^{a}$}
Stephen Yang,\textit{$^{b}$}
and Ambuj Tewari\textit{$^{a,b}$}
}

\end{center}


\renewcommand*\rmdefault{bch}
\normalfont
\upshape
\rmfamily


\footnotetext{
\textit{
$^{a}$~Department of Statistics, University of Michigan,
Ann Arbor, Michigan, USA.
E-mail: yikunhan@umich.edu; ywee@umich.edu;
nmankodi@umich.edu; tewaria@umich.edu
}
}

\footnotetext{
\textit{
$^{b}$~Computer Science and Engineering Division,
University of Michigan, Ann Arbor, Michigan, USA.
E-mail: yfming@umich.edu
}
}

\footnotetext{
\ddag~These authors contributed equally to this work.
}


\vspace{1.0cm}

\begin{center}
\small

\captionof{table}{
Comparison of the 84M- and 164M-parameter Uni-Mol2 models on the
GS-LF test set. Each cell reports the point estimate with its 95\%
bootstrap confidence interval based on 1000 resamples of the stratified
test set. Label-wise thresholds were learned separately from the
corresponding GS-LF out-of-fold predictions and applied to the held-out
test set.
}

\label{tab:model-size-comparison}

\vspace{0.4em}

\begin{tabular*}{\textwidth}{
    @{\extracolsep{\fill}}lccccc
}

\toprule

Model
& Macro AUROC
& Macro AUPRC
& Macro F1
& Macro Precision
& Macro Recall \\

\midrule

Uni-Mol2 (FT, 84M)
& \textbf{0.8990}
& \textbf{0.4077}
& \textbf{0.3991}
& \textbf{0.4288}
& 0.4291 \\

&
{\scriptsize [0.8923, 0.9062]}
&
{\scriptsize [0.4029, 0.4444]}
&
{\scriptsize [0.3720, 0.4119]}
&
{\scriptsize [0.3949, 0.4488]}
&
{\scriptsize [0.4051, 0.4517]} \\

\addlinespace

Uni-Mol2 (FT, 164M)
& 0.8975
& 0.4042
& 0.3954
& 0.4111
& \textbf{0.4302} \\

&
{\scriptsize [0.8906, 0.9048]}
&
{\scriptsize [0.3992, 0.4407]}
&
{\scriptsize [0.3683, 0.4074]}
&
{\scriptsize [0.3863, 0.4336]}
&
{\scriptsize [0.4061, 0.4517]} \\

\bottomrule

\end{tabular*}

\end{center}


\clearpage
\begingroup
\small
\setlength{\tabcolsep}{3pt}
\newcommand{\pairedci}[2]{{\scriptsize $[#1,\,#2]$}}

\begin{center}
\captionof{table}{
Paired performance differences between Uni-Mol2 (FT) and OpenPOM.
Each point estimate is followed by its 95\% percentile confidence interval from 5000 paired bootstrap resamples. Bold estimates indicate intervals that exclude zero. In Panel A, differences are computed as Uni-Mol2 (FT) minus OpenPOM. In Panel B, RMSE reduction is computed as OpenPOM minus Uni-Mol2 (FT), and Pearson correlation gain as Uni-Mol2 (FT) minus OpenPOM. Positive values favor Uni-Mol2 (FT) in both panels.
}
\label{tab:paired-comparisons}

\vspace{0.4em}

\begin{tabular*}{\textwidth}{@{\extracolsep{\fill}}lccccc@{}}
\multicolumn{6}{c}{\textbf{Panel A. Classification tasks}} \\
\toprule
Dataset & AUROC & AUPRC & F1 & Precision & Recall \\
\midrule

GS-LF
& \textbf{0.0107} & \textbf{0.0329} & \textbf{0.0243}
& \textbf{0.0228} & \textbf{0.0243} \\
& \pairedci{0.0071}{0.0144} & \pairedci{0.0214}{0.0414}
& \pairedci{0.0108}{0.0373} & \pairedci{0.0006}{0.0435}
& \pairedci{0.0085}{0.0398} \\
\addlinespace

Enantiomer subset
& 0.0272 & \textbf{0.1187} & 0.0127 & 0.0115 & 0.0048 \\
& \pairedci{-0.0302}{0.0744} & \pairedci{0.0574}{0.1853}
& \pairedci{-0.0008}{0.0158} & \pairedci{0.0000}{0.0161}
& \pairedci{-0.0143}{0.0095} \\
\addlinespace

Zhang (complete)
& 0.0063 & \textbf{0.0408} & \textbf{0.0264}
& \textbf{0.0209} & \textbf{0.0465} \\
& \pairedci{-0.0002}{0.0122} & \pairedci{0.0223}{0.0613}
& \pairedci{0.0112}{0.0431} & \pairedci{0.0038}{0.0385}
& \pairedci{0.0242}{0.0682} \\
\addlinespace

Zhang (non-overlap)
& 0.0038 & \textbf{0.0672} & \textbf{0.0244}
& 0.0189 & \textbf{0.0430} \\
& \pairedci{-0.0079}{0.0115} & \pairedci{0.0260}{0.0799}
& \pairedci{0.0025}{0.0415} & \pairedci{-0.0070}{0.0370}
& \pairedci{0.0048}{0.0646} \\
\addlinespace

Odor Detection (complete)
& 0.0133 & $\mathbf{-0.0435}$ & \textbf{0.1194}
& $-0.0663$ & \textbf{0.1877} \\
& \pairedci{-0.0002}{0.0269} & \pairedci{-0.0785}{-0.0049}
& \pairedci{0.0643}{0.1771} & \pairedci{-0.1363}{0.0055}
& \pairedci{0.1329}{0.2420} \\
\addlinespace

Odor Detection (canonical non-overlap)
& 0.0170 & $-0.0395$ & \textbf{0.1221}
& $-0.0579$ & \textbf{0.1957} \\
& \pairedci{-0.0013}{0.0355} & \pairedci{-0.0750}{0.0009}
& \pairedci{0.0647}{0.1829} & \pairedci{-0.1327}{0.0182}
& \pairedci{0.1387}{0.2536} \\
\bottomrule
\end{tabular*}
\end{center}


\begin{center}

\vspace{0.4em}

\begin{tabular*}{\textwidth}{@{\extracolsep{\fill}}llcc@{}}
\multicolumn{4}{c}{\textbf{Panel B. Mixture prediction}} \\
\toprule
Dataset (test pairs) & Regressor comparison & RMSE reduction & Pearson gain \\
\midrule

Combined (548) & ElasticNet vs ElasticNet & \textbf{0.0034} & \textbf{0.0528} \\
& & \pairedci{0.0020}{0.0048} & \pairedci{0.0289}{0.0804} \\

& RF vs RF & 0.0017 & 0.0164 \\
& & \pairedci{-0.0004}{0.0037} & \pairedci{-0.0063}{0.0386} \\

& GBT vs GBT & \textbf{0.0041} & 0.0359 \\
& & \pairedci{0.0004}{0.0077} & \pairedci{-0.0008}{0.0730} \\

& OpenPOM RF vs Uni-Mol2 GBT & 0.0030 & 0.0215 \\
& & \pairedci{-0.0003}{0.0062} & \pairedci{-0.0109}{0.0540} \\
\midrule

Bushdid (260) & ElasticNet vs ElasticNet & \textbf{0.0025} & 0.0307 \\
& & \pairedci{0.0013}{0.0037} & \pairedci{-0.0111}{0.0693} \\

& RF vs RF & $-0.0017$ & $-0.0147$ \\
& & \pairedci{-0.0048}{0.0013} & \pairedci{-0.0455}{0.0156} \\

& GBT vs GBT & $-0.0016$ & $-0.0177$ \\
& & \pairedci{-0.0066}{0.0035} & \pairedci{-0.0666}{0.0333} \\
\midrule

Snitz 1 (147) & ElasticNet vs ElasticNet & 0.0020 & 0.0161 \\
& & \pairedci{-0.0034}{0.0067} & \pairedci{-0.0181}{0.0591} \\

& RF vs RF & \textbf{0.0074} & \textbf{0.0924} \\
& & \pairedci{0.0040}{0.0108} & \pairedci{0.0394}{0.1514} \\

& GBT vs GBT & \textbf{0.0141} & \textbf{0.1539} \\
& & \pairedci{0.0073}{0.0206} & \pairedci{0.0677}{0.2397} \\
\midrule

Snitz 2 (91) & ElasticNet vs ElasticNet & \textbf{0.0103} & \textbf{0.2146} \\
& & \pairedci{0.0063}{0.0145} & \pairedci{0.0894}{0.3511} \\

& RF vs RF & \textbf{0.0061} & 0.0691 \\
& & \pairedci{0.0007}{0.0116} & \pairedci{-0.0337}{0.1805} \\

& GBT vs GBT & \textbf{0.0098} & \textbf{0.1399} \\
& & \pairedci{0.0016}{0.0184} & \pairedci{0.0131}{0.2817} \\
\midrule

Ravia (50) & ElasticNet vs ElasticNet & \textbf{0.0027} & 0.0222 \\
& & \pairedci{0.0007}{0.0050} & \pairedci{-0.0455}{0.0954} \\

& RF vs RF & 0.0013 & 0.0131 \\
& & \pairedci{-0.0036}{0.0064} & \pairedci{-0.0712}{0.1018} \\

& GBT vs GBT & 0.0064 & 0.1122 \\
& & \pairedci{-0.0075}{0.0187} & \pairedci{-0.0842}{0.3049} \\
\bottomrule
\end{tabular*}
\end{center}

\endgroup

\clearpage
\begingroup

\begin{center}
\small

\captionof{table}{
Performance on the Zhang and odor detection datasets by chemical distance
from the GS-LF training set. For each test molecule, distance was defined
as one minus its maximum fingerprint Tanimoto similarity to any GS-LF
training molecule. Each distance threshold defines a subset of the
corresponding non-overlapping test set. Entries report AUROC / AUPRC.
For Zhang, valid labels are those included in the macro averages. N denotes the number of test molecules in each subset.
}
\label{tab:chemical-distance}

\vspace{0.4em}

\begin{tabular*}{\textwidth}{@{\extracolsep{\fill}}lcccc@{}}
\toprule
Subset & $N$ & Valid labels & OpenPOM & Uni-Mol2 (FT) \\
\midrule

\multicolumn{5}{l}{\textbf{Zhang}} \\
Original non-overlap
& 506 & 105 & 0.8937 / 0.2937 & 0.8975 / 0.3609 \\
Distance $> 0$ (Q25)
& 356 & 97 & 0.8724 / 0.3030 & 0.8716 / 0.3535 \\
Distance $> 0.3590$ (Q50)
& 252 & 82 & 0.8444 / 0.2773 & 0.8417 / 0.3194 \\
Distance $> 0.5455$ (Q75)
& 126 & 45 & 0.7666 / 0.2594 & 0.7731 / 0.2622 \\

\midrule

\multicolumn{5}{l}{\textbf{Odor Detection}} \\
Canonical non-overlap
& 1,378 & -- & 0.7727 / 0.4423 & 0.7897 / 0.4028 \\
Distance $> 0$
& 1,317 & -- & 0.7628 / 0.4341 & 0.7808 / 0.3948 \\
Distance $> 0.4615$ (Q25)
& 1,022 & -- & 0.7302 / 0.4319 & 0.7483 / 0.3919 \\
Distance $> 0.6154$ (Q50)
& 674 & -- & 0.6527 / 0.4544 & 0.6728 / 0.4113 \\
Distance $> 0.7333$ (Q75)
& 341 & -- & 0.6360 / 0.5202 & 0.6248 / 0.4404 \\
\bottomrule
\end{tabular*}

\end{center}
\endgroup

\clearpage
\begingroup

\begin{center}
\small
\setlength{\tabcolsep}{3pt}

\captionof{table}{
Leave-one-dataset-out evaluation of odor mixture prediction.
For each held-out dataset, regressors were trained on the other three
datasets. Parentheses give the number of test pairs. Model performance
is reported as RMSE / Pearson correlation. RMSE reduction is computed
as OpenPOM minus Uni-Mol2 (FT), and Pearson gain as Uni-Mol2 (FT)
minus OpenPOM. Positive differences favor Uni-Mol2 (FT). Each
difference is followed by its 95\% percentile confidence interval
from 5000 paired bootstrap resamples. Bold differences have
intervals that exclude zero.
}
\label{tab:mixture-held-out}

\vspace{0.4em}

\begin{tabular*}{\textwidth}{@{\extracolsep{\fill}}llcccc@{}}
\toprule
Held-out dataset & Regressor & OpenPOM & Uni-Mol2 (FT)
& RMSE reduction & Pearson gain \\
\midrule

Bushdid (260) & ElasticNet & 0.169 / 0.379 & 0.171 / 0.387
& $\mathbf{-0.0017}$ & 0.0075 \\
& & & &
{\scriptsize $[-0.0029, -0.0005]$} &
{\scriptsize $[-0.0392, 0.0528]$} \\

& RF & 0.161 / 0.453 & 0.159 / 0.485
& \textbf{0.0024} & 0.0327 \\
& & & &
{\scriptsize $[0.0002, 0.0046]$} &
{\scriptsize $[-0.0067, 0.0748]$} \\

& GBT & 0.174 / 0.244 & 0.161 / 0.413
& \textbf{0.0123} & \textbf{0.1687} \\
& & & &
{\scriptsize $[0.0063, 0.0184]$} &
{\scriptsize $[0.0829, 0.2596]$} \\

\midrule

Snitz 1 (147) & ElasticNet & 0.135 / 0.390 & 0.128 / 0.425
& \textbf{0.0074} & 0.0357 \\
& & & &
{\scriptsize $[0.0010, 0.0137]$} &
{\scriptsize $[-0.0166, 0.0929]$} \\

& RF & 0.129 / 0.240 & 0.127 / 0.291
& 0.0022 & 0.0517 \\
& & & &
{\scriptsize $[-0.0019, 0.0064]$} &
{\scriptsize $[-0.0630, 0.1700]$} \\

& GBT & 0.132 / 0.173 & 0.124 / 0.336
& \textbf{0.0082} & \textbf{0.1628} \\
& & & &
{\scriptsize $[0.0026, 0.0138]$} &
{\scriptsize $[0.0449, 0.2906]$} \\

\midrule

Snitz 2 (91) & ElasticNet & 0.170 / 0.265 & 0.145 / 0.258
& \textbf{0.0246} & $-0.0066$ \\
& & & &
{\scriptsize $[0.0145, 0.0364]$} &
{\scriptsize $[-0.1035, 0.0946]$} \\

& RF & 0.157 / 0.399 & 0.152 / 0.499
& 0.0047 & 0.0998 \\
& & & &
{\scriptsize $[-0.0004, 0.0098]$} &
{\scriptsize $[-0.0330, 0.2361]$} \\

& GBT & 0.158 / 0.258 & 0.156 / 0.500
& 0.0026 & \textbf{0.2417} \\
& & & &
{\scriptsize $[-0.0051, 0.0107]$} &
{\scriptsize $[0.0373, 0.4692]$} \\

\midrule

Ravia (50) & ElasticNet & 0.137 / 0.363 & 0.139 / 0.374
& $-0.0017$ & 0.0112 \\
& & & &
{\scriptsize $[-0.0042, 0.0004]$} &
{\scriptsize $[-0.0305, 0.0593]$} \\

& RF & 0.133 / 0.395 & 0.131 / 0.445
& 0.0025 & 0.0496 \\
& & & &
{\scriptsize $[-0.0040, 0.0088]$} &
{\scriptsize $[-0.0595, 0.1570]$} \\

& GBT & 0.132 / 0.413 & 0.132 / 0.434
& 0.0006 & 0.0204 \\
& & & &
{\scriptsize $[-0.0155, 0.0144]$} &
{\scriptsize $[-0.1960, 0.2067]$} \\

\bottomrule
\end{tabular*}

\end{center}
\endgroup